\documentclass[sigconf,screen,nonacm]{acmart}
\renewcommand\footnotetextcopyrightpermission[1]{}
\AtBeginDocument{%
  }

\usepackage{graphicx}
\usepackage{booktabs}
\usepackage{multirow}
\usepackage{subcaption}
\usepackage{enumitem}
\usepackage{xspace}
\usepackage{graphicx}
\usepackage{amssymb}
\usepackage[table]{xcolor}
\usepackage{algorithm}
\usepackage{algpseudocode}
\usepackage[capitalise,noabbrev]{cleveref}
\newcommand{\etc}{\textit{etc.}\xspace}

\begin{document}

\title{SubFlow: Sub-mode Conditioned Flow Matching for Diverse One-Step Generation}

\author{Yexiong Lin}
\affiliation{%
  \institution{Sydney AI Centre, The University of Sydney}
  \city{Sydney}
  \state{NSW}
  \country{Australia}
}

\author{Jia Shi}
\affiliation{%
  \institution{School of Artificial Intelligence, Xidian University}
  \city{Xi'an}
  \state{Shaanxi}
  \country{China}
}

\author{Shanshan Ye}
\affiliation{%
  \institution{Australian Artificial Intelligence Institute, University of Technology Sydney}
  \city{Sydney}
  \state{NSW}
  \country{Australia}
}

\author{Wanyu Wang}
\affiliation{%
  \institution{Department of Information Systems, City University of Hong Kong}
  \city{Hong Kong}
  \country{China}
}

\author{Yu Yao}
\affiliation{%
  \institution{Sydney AI Centre, The University of Sydney}
  \city{Sydney}
  \state{NSW}
  \country{Australia}
}

\author{Tongliang Liu}
\affiliation{%
  \institution{Sydney AI Centre, The University of Sydney}
  \city{Sydney}
  \state{NSW}
  \country{Australia}
}

\renewcommand{\shortauthors}{Lin et al.}

\begin{abstract}
Flow matching has emerged as a powerful generative framework, with recent few-step methods achieving remarkable inference acceleration.
However, we identify a critical yet overlooked limitation: these models suffer from severe \textit{diversity degradation}, concentrating samples on dominant modes while neglecting rare but valid variations of the target distribution.
We trace this degradation to \textit{averaging distortion}: when trained with MSE objectives, class-conditional flows learn a frequency-weighted mean over intra-class sub-modes, causing the model to over-represent high-density modes while systematically neglecting low-density ones.
To address this, we propose SubFlow, Sub-mode Conditioned Flow Matching, which eliminates averaging distortion by decomposing each class into fine-grained sub-modes via semantic clustering and conditioning the flow on sub-mode indices.
Each conditioned sub-distribution is approximately unimodal, so the learned flow accurately targets individual modes with no averaging distortion, restoring full mode coverage in a single inference step.
Crucially, SubFlow is entirely \textit{plug-and-play}: it integrates seamlessly into existing one-step models such as MeanFlow and Shortcut Models without any architectural modifications.
Extensive experiments on ImageNet-256 demonstrate that SubFlow yields substantial gains in generation diversity (Recall) while maintaining competitive image quality (FID), confirming its broad applicability across different one-step generation frameworks. Project page: \url{https://yexionglin.github.io/subflow}.
\end{abstract}


\maketitle

\section{Introduction}
\label{sec:intro}

Recent advances in generative models, particularly diffusion models~\cite{ho2020denoising,song2020score,song2019generative,karras2022elucidating} and flow matching~\cite{lipman2022flow,liu2022flow,tong2023improving}, have achieved remarkable success in synthesizing high-quality images, audio, and other modalities~\cite{zheng2025aligning,hong2025adlift}. However, these models typically require a large number of sampling steps (50-1000 steps) to produce high-fidelity samples, resulting in prohibitive inference costs that limit their deployment in real-world scenarios~\cite{qu2025spatialvla,qu2025eo,Zhou_2024_CVPR,zhou2026drivinggen,zhou2026drivedreamerpolicy}. This has motivated extensive research on few-step generation methods~\cite{song2023consistency,song2023improved,lu2025simplifying,salimans2022progressive,liu2024instaflow,lin2025beyond}, which aim to accelerate inference by 10x-100x while maintaining generation quality. These approaches include distillation from pre-trained models~\cite{salimans2022progressive,yin2024one,luo2023diff}, consistency training that enforces self-consistency along trajectories~\cite{song2023consistency,song2023improved,lu2025simplifying,geng2024consistency,kim2024consistency}, and MeanFlow objectives that explicitly learn the average-velocity vector field for one-step transport~\cite{geng2025mean}.

\begin{figure*}[t]
  \centering
  \begin{subfigure}[b]{0.32\textwidth}
    \includegraphics[width=\textwidth]{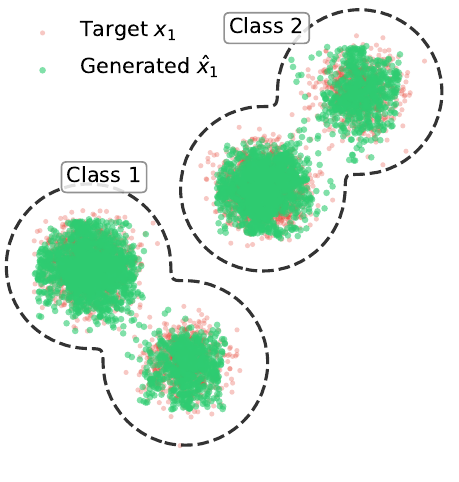}
    \caption{Traditional FM (100 steps)}
    \label{fig:toy_trad}
  \end{subfigure}
  \hfill
  \begin{subfigure}[b]{0.32\textwidth}
    \includegraphics[width=\textwidth]{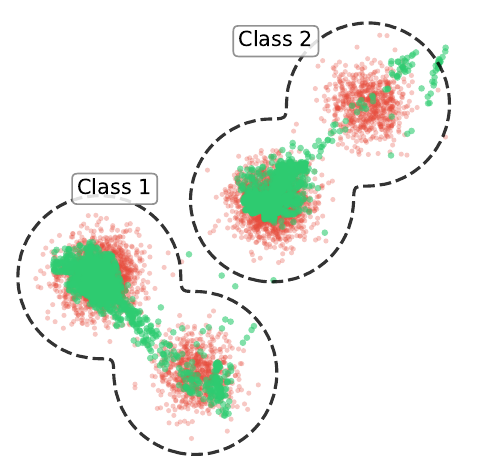}
    \caption{MeanFlow (1 step)}
    \label{fig:toy_baseline}
  \end{subfigure}
  \hfill
  \begin{subfigure}[b]{0.32\textwidth}
    \includegraphics[width=\textwidth]{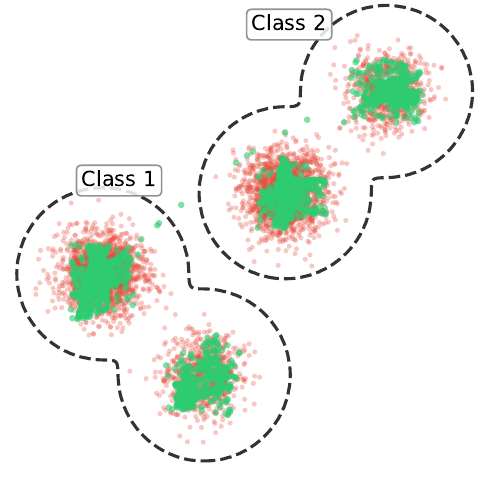}
    \caption{SubFlow (1 step)}
    \label{fig:toy_ours}
  \end{subfigure}
  \caption{Toy experiment on a 4-peak Gaussian mixture target with imbalanced subclusters (majority weight 0.35, minority weight 0.15 per class). \textbf{(a)} Traditional flow matching (100 inference steps) covers all modes. \textbf{(b)} MeanFlow suffers from severe mode collapse, concentrating only on dominant subclusters and losing diversity. \textbf{(c)} In contrast, our SubFlow successfully restores full mode coverage in a single step by decomposing each class into fine-grained sub-modes and conditioning the flow on sub-mode indices.}
  \label{fig:toy}
\end{figure*}

Despite their impressive speed improvements, existing few-step generation methods suffer from a critical limitation: significant \textit{diversity degradation}. Specifically, this refers to a systematic loss of distributional coverage, where the model heavily biases towards common data patterns and ignores rare, minority variations. Empirically, we observe that reducing the sampling budget actively induces this phenomenon. As illustrated in \cref{fig:toy_baseline}, the generated distribution undergoes a severe shift, concentrating almost entirely around dominant modes and failing to capture minor, low-density modes.
Such pronounced \textit{mode collapse}, i.e., the failure of the model to generate samples from all regions of the target distribution, severely restricts the practical applications of few-step generation models, as diversity is important for downstream applications such as creative content generation, data augmentation, and preventing bias against underrepresented groups. Consequently, understanding and mitigating this diversity loss remains a pressing open challenge in the field~\cite{gandikota2025distilling}.

We identify the fundamental cause of this diversity loss as the \textit{averaging distortion} inherent in the class-conditional flow matching objective. This distortion occurs when the model is forced to compromise between conflicting target directions, transporting samples toward the dominant mode instead of covering all distinct sub-modes. Specifically, by minimizing the mean squared error over training pairs, the optimal vector field $v_\theta(x_t, t, c)$ converges to the conditional mean velocity $\mathbb{E}[x_1 - x_0 \mid x_t, t, c]$. When the target distribution within class $c$ contains multiple sub-modes with unequal densities, this conditional mean acts as a frequency-weighted average biased toward high-density regions. As a result, the model introduces a \textit{dominant-mode bias}, implicitly favoring dominant sub-modes while abandoning low-density ones.
One might expect multi-step ODE integration to mitigate this bias through iterative refinement. However, if the learned vector field is itself biased toward dominant sub-modes, additional integration steps may only trace this biased field more accurately. In this case, reducing discretization error can improve sample fidelity, but may be insufficient to recover the missing diversity. Our experiments support this view: increasing the number of integration steps could improve FID, while Recall remains largely unchanged or even decreases slightly (\cref{sec:ablation}).
Generated samples thus systematically concentrate on dominant sub-modes, resulting in severe \textit{mode collapse}. Our toy experiments explicitly validate this mechanism. As shown in \cref{fig:toy_baseline}, the single-step baseline not only collapses onto dominant modes but also produces samples displaced into the low-density space between clusters.

To directly eliminate this dominant-mode bias, we propose Sub-mode Conditioned Flow Matching (SubFlow). The key insight is that dominant-mode bias arises because the conditional mean velocity $\mathbb{E}[x_1 - x_0 \mid x_t, t, c]$ must average over all sub-modes within class $c$. By further conditioning on a sub-mode index $k$, each sub-distribution $(c, k)$ becomes approximately unimodal, and the conditional mean $\mathbb{E}[x_1 - x_0 \mid x_t, t, c, k]$ accurately points to a specific mode with no averaging distortion. 
Specifically, we first leverage a pre-trained vision foundation model (i.e., DINOv3 \cite{simeoni2025dinov3}) to extract semantic features and partition the target data manifold into $K$ distinct sub-manifolds, effectively refining each broad class into fine-grained sub-modes. We then optimize a sub-mode-conditioned vector field $v_\theta(x_t, t, c, k)$ that explicitly takes both the original class label $c$ and the specific sub-mode index $k \in \{1, \dots, K\}$ as conditions. 
This fine-grained supervision ensures that each index $k$ defines a unique, unambiguous ODE trajectory, enabling the model to learn $K$ independent conditional flows that fairly preserve every local mode within the class. During inference, diverse outputs are generated by sampling the sub-mode indices according to their empirical prior distribution. Crucially, our method is entirely \textit{plug-and-play}. It can be seamlessly integrated into state-of-the-art one-step models, such as MeanFlow~\cite{geng2025mean} and Shortcut Models~\cite{frans2025one}, without requiring any architectural modifications or alterations to the underlying ODE solver. This inherent flexibility renders our approach strictly orthogonal and complementary to existing acceleration paradigms.

Our main contributions are summarized as follows:
\begin{itemize}[leftmargin=*]
    \item \textbf{Problem Identification}: We identify \textit{averaging distortion}, i.e., the tendency of MSE-trained class-conditional flows to learn a frequency-weighted mean over intra-class sub-modes, as the root cause of diversity degradation in few-step generation.
    \item \textbf{Method Innovation}: We propose Sub-mode Conditioned Flow Matching (SubFlow), which eliminates averaging distortion through sub-manifold decomposition and conditioning on sub-mode indices.
    \item \textbf{Plug-and-Play Design}: Our method can be seamlessly integrated into existing one-step models (MeanFlow, Shortcut Models, \etc) without architectural changes, demonstrating broad applicability.
    \item \textbf{Comprehensive Validation}: We extensively evaluate SubFlow on the challenging ImageNet-256 benchmark by integrating it with three state-of-the-art one-step generation frameworks: MeanFlow~\cite{geng2025mean}, Shortcut Models~\cite{frans2025one}, and SoFlow \cite{luo2026soflow}. Our empirical results demonstrate that SubFlow drastically mitigates diversity degradation, yielding substantial gains in Recall (e.g., 43.45\%$\to$48.84\% for MeanFlow) while maintaining competitive overall image fidelity (FID). Consistent performance enhancements across all three baselines, supported by detailed ablation studies, confirm the robustness and broad applicability of our approach.
\end{itemize}

\section{Related Work}
\label{sec:related}

\paragraph{Diffusion Models and Flow Matching.}
Diffusion models~\cite{ho2020denoising,song2020score,song2019generative,dhariwal2021diffusion,karras2022elucidating} are a dominant paradigm for generative modeling and have achieved strong results in image, audio, and video synthesis~\cite{zheng2026vii}. They learn to reverse a forward noising process through iterative denoising. Latent diffusion models~\cite{rombach2022high} further improve efficiency by operating in a compressed latent space. Flow matching~\cite{lipman2022flow,liu2022flow,tong2023improving} provides an alternative formulation that directly learns a vector field to transport a source distribution to a target distribution via Continuous Normalizing Flows (CNFs)~\cite{chen2018neural}. Compared with standard diffusion training, conditional flow matching can leverage more structured couplings (e.g., optimal transport couplings)~\cite{tong2023improving,albergo2022building}, often yielding straighter trajectories and improved training efficiency. Recent large-scale text-to-image systems (e.g., Stable Diffusion 3) have further demonstrated the practicality and scalability of flow-matching-style formulations for high-quality generation~\cite{esser2024scaling}. Our work builds on conditional flow matching~\cite{tong2023improving} and focuses on its diversity limitation in the few-step regime.

\paragraph{Few-Step Generation.}
Reducing sampling steps while maintaining generation quality is a central challenge in modern generative modeling. One line of work is \textit{distillation-based acceleration}: progressive distillation~\cite{salimans2022progressive} compresses multiple teacher denoising steps into fewer student steps; InstaFlow~\cite{liu2024instaflow} distills multistep flow models for one-step generation; distribution matching distillation~\cite{yin2024one} and score-based distillation~\cite{luo2024one,zhou2024score,luo2023diff,chen2025pi} enable one-step generation by matching the score or distribution of a pre-trained teacher; and guided distillation~\cite{meng2023distillation} further extends these ideas to conditional generation. Another major line of work is \textit{consistency-based methods}: consistency models~\cite{song2023consistency} enforce self-consistency along probability flow ODE trajectories, with subsequent variants improving training stability and scalability~\cite{song2023improved,lu2025simplifying,geng2024consistency,kim2024consistency}. A third line explores \textit{direct few-step training} without relying on a separately distilled teacher: Shortcut Models~\cite{frans2025one} use self-consistency constraints to support few-step generation; MeanFlow~\cite{geng2025mean} estimates mean trajectory velocities for one-step generation; and inductive moment matching~\cite{zhou2025inductive} achieves one-step generation via moment-based objectives.
Recent work has also extended standard first-order flow or rectified-flow dynamics in several directions. Cao et al.~\cite{cao2025force} proposed Force Matching, a physics-inspired variant that imposes relativistic velocity constraints for more stable sampling, while Zhang et al.~\cite{zhang2025towards} introduced Hierarchical Rectified Flow, which hierarchically couples ODEs over location, velocity, and acceleration to model richer stochastic dynamics. Su et al.~\cite{su2025high} further developed a unified high-order flow matching framework with arbitrary-order trajectory information and sharp statistical guarantees.
While these methods substantially improve sampling speed, prior work has paid limited attention to the systematic degradation of mode coverage and diversity in the few-step setting.

\paragraph{Diversity Degradation in Generative Models.}
The quality--diversity trade-off is a recurring theme across generative paradigms. In GANs, mode collapse has been extensively studied~\cite{goodfellow2014generative,metz2016unrolled,srivastava2017veegan}, and the truncation trick explicitly trades sample variety for fidelity~\cite{brock2018large}. In diffusion models, strong classifier-free guidance (CFG) sharpens image quality at the expense of mode coverage~\cite{ho2022classifier,kynkaanniemi2024applying,xiang2026safety}. A common thread in prior work is to attribute this degradation to \emph{inference-time} design choices and to propose inference-time remedies accordingly, such as condition-annealed sampling~\cite{sadat2024cads}. In contrast, the diversity loss we study is rooted in the \emph{training objective} itself: when the target distribution is multi-modal, class-conditional MSE regression produces a velocity field inherently biased toward dominant sub-modes, regardless of the number of sampling steps.

Meanwhile, clustering and mixture decompositions have been incorporated into generative modeling in various forms, including latent quantization~\cite{van2017neural} and expert routing~\cite{shazeer2017outrageously}, but these techniques target model capacity or representation structure rather than the diversity of generated samples. To the best of our knowledge, existing flow matching methods have not explicitly modeled the multi-modal structure within each class. Our work takes a step in this direction: we cluster each class into finer sub-modes and use the resulting partition as an augmented conditioning signal during training, making each regression target locally unimodal and thereby reducing the averaging distortion.

\begin{figure*}[t]
  \centering
  \includegraphics[width=0.82\textwidth,trim=0 30 130 0,clip]{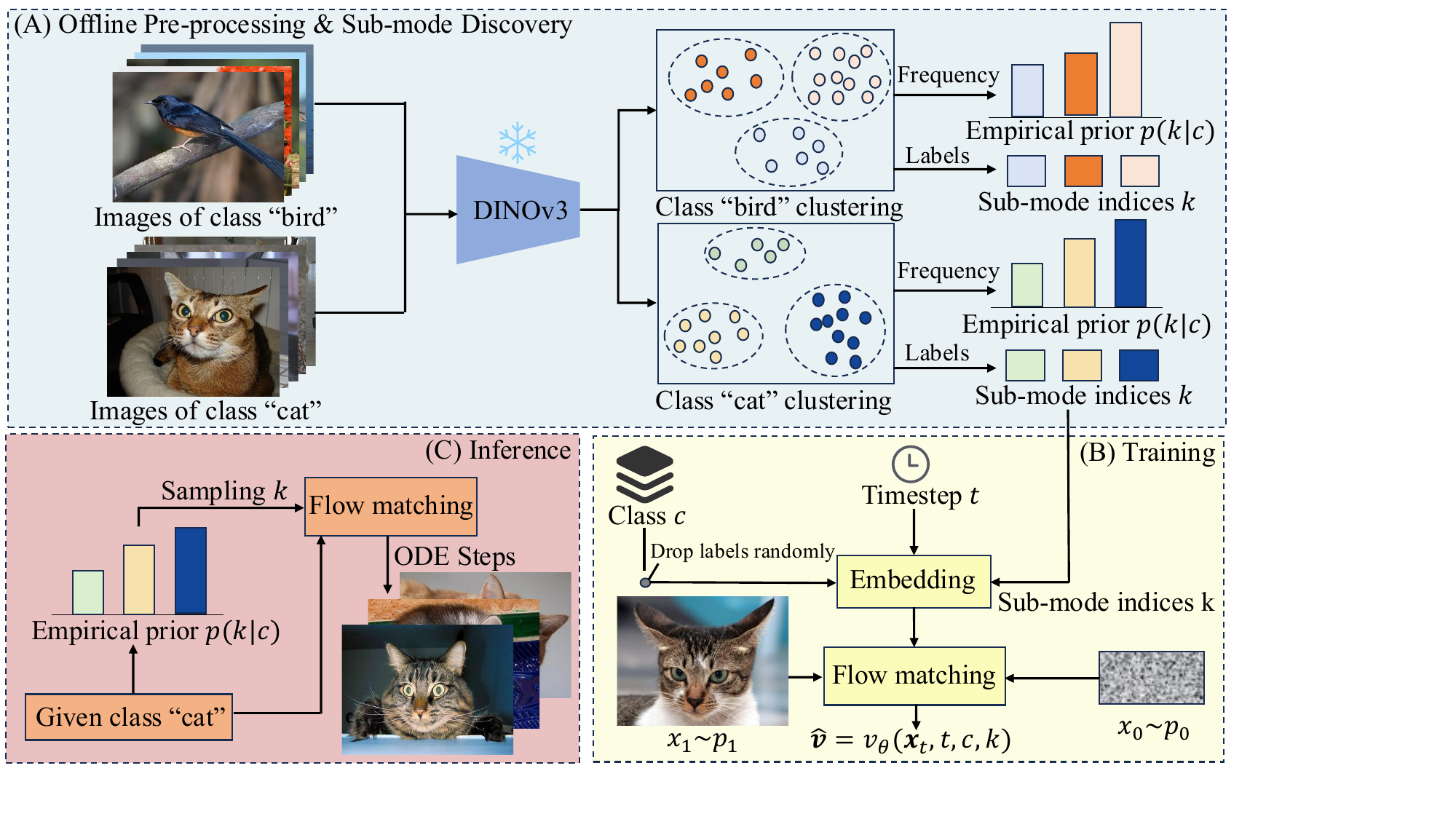}
  \vspace{-10px}
  \caption{Overview of SubFlow.
  \textbf{(a) Offline pre-processing}: semantic features are extracted from the training images and clustered within each class, yielding sub-mode assignments $\{k_i\}$ and empirical priors $p(k\mid c)$.
  \textbf{(b) Training}: the vector field $v_\theta(x_t, t, c, k)$ is optimized with the SubFlow objective. For classifier-free guidance, only the class label $c$ is randomly dropped, while the sub-mode index $k$ is always retained.
  \textbf{(c) Inference}: given a target class $c$, a sub-mode index $k$ is sampled from $p(k\mid c)$, and the conditioned vector field generates a sample associated with the selected sub-mode.}
  \label{fig:method}
\end{figure*}

\section{Method}
\label{sec:method}

\subsection{Preliminaries: Flow Matching}
\label{sec:prelim}

Let $p_0$ denote a tractable source distribution (e.g., a standard Gaussian $\mathcal{N}(0, I)$), and let $p_1$ denote the target data distribution. Flow matching (FM)~\cite{lipman2022flow,tong2023improving} learns a time-dependent vector field $v_\theta: \mathbb{R}^d \times [0,1] \to \mathbb{R}^d$ whose associated ordinary differential equation (ODE) induces a flow that transports the source distribution $p_0$ to the target distribution $p_1$:
\begin{equation}
  \frac{dx_t}{dt} = v_\theta(x_t, t), \quad x_0 \sim p_0.
  \label{eq:ode}
\end{equation}

Let $p_t$ denote the distribution of $x_t$ at time $t$. The target vector field $u_t(x)$ defines the probability path $\{p_t\}_{t \in [0,1]}$ connecting $p_0$ and $p_1$, and the pair $(p_t, u_t)$ should satisfy the continuity equation
\begin{equation}
  \partial_t p_t(x) + \nabla \cdot \bigl( p_t(x) u_t(x) \bigr) = 0.
  \label{eq:continuity}
\end{equation}
The ideal FM objective regresses $v_\theta$ to this marginal velocity field:
\begin{equation}
  \mathcal{L}_{\mathrm{FM}} = \mathbb{E}_{t \sim \mathcal{U}[0,1],\, x \sim p_t}
  \bigl\| v_\theta(x, t) - u_t(x) \bigr\|^2.
  \label{eq:fm}
\end{equation}
However, directly supervising $u_t(x)$ is usually intractable, since the marginal probability path $p_t$ and its associated velocity field are generally not available in practice.

To bypass this difficulty, Lipman et al.~\cite{lipman2022flow} formulate the training via \emph{Conditional Flow Matching} (CFM). Instead of supervising the marginal field $u_t$, CFM defines a family of conditional probability paths $\{p_t(\cdot \mid x_1)\}$ indexed by data samples $x_1 \sim p_1$, together with corresponding conditional vector fields $u_t(\cdot \mid x_1)$.

For the commonly used linear conditional path, one first samples $x_0 \sim p_0$ and $x_1 \sim p_1$, and then defines
\begin{equation}
  x_t = (1-t)x_0 + t x_1,
  \qquad
  u_t(x_t \mid x_0, x_1) = x_1 - x_0.
  \label{eq:linear_path}
\end{equation}
The corresponding CFM objective minimizes
\begin{equation}
  \mathcal{L}_{\mathrm{CFM}} = \mathbb{E}_{t,\, x_0,\, x_1} \bigl\| v_\theta(x_t, t) - (x_1 - x_0) \bigr\|^2,
  \label{eq:cfm}
\end{equation}
where $t \sim \mathcal{U}[0,1]$, $x_0 \sim p_0$, and $x_1 \sim p_1$.
A key theoretical result in~\cite{lipman2022flow} is that, under a valid conditional probability path construction, CFM provides an unbiased estimator of the corresponding FM gradient.
In particular, for the linear interpolation path, the marginal velocity field satisfies
\begin{equation}
  u_t(x) = \mathbb{E}[x_1 - x_0 \mid x_t = x].
  \label{eq:marginal_velocity}
\end{equation}
Therefore, optimizing Eq. \eqref{eq:cfm} provides a tractable way to learn the same target marginal vector field as Eq. \eqref{eq:fm}, while requiring only samples from $(x_0, x_1)$ and an analytically specified conditional path. At inference, a sample is generated by integrating Eq. \eqref{eq:ode} from $x_0 \sim p_0$ to $t=1$.

\subsection{Averaging Distortion in Class-Conditional Flow Matching}
\label{sec:avgdist}

We now consider class-conditional generation with semantic label $c$. Note that this class condition is distinct from the ``conditional'' in Conditional Flow Matching, which refers to conditioning on path variables such as $x_1$. In the class-conditional setting, the training objective becomes
\begin{equation}
  \mathcal{L}_{\mathrm{cCFM}} = \mathbb{E}_{t,\, x_0,\, (x_1,c)} \bigl\| v_\theta(x_t, t, c) - (x_1 - x_0) \bigr\|^2.
  \label{eq:ccfm}
\end{equation}
In this formulation, the model is required to learn a single vector field for all samples belonging to class $c$. However, when the target distribution within class $c$ is multi-modal, samples from different sub-modes may induce distinct transport directions under the same class condition. This creates an intrinsic ambiguity for learning a single class-conditional vector field.

\paragraph{Optimal vector field.}
Under the squared-error objective, the population-optimal class-conditional vector field under Eq. \eqref{eq:ccfm} is the conditional mean velocity:
\begin{equation}
  v_\theta^*(x_t, t, c) = \mathbb{E}[x_1 - x_0 \mid x_t, t, c].
  \label{eq:optimal}
\end{equation}
For any fixed $(x_t, t, c)$, the expected squared error $\mathbb{E}[\|v - (x_1-x_0)\|^2]$ is minimized by the conditional mean. Consequently, the learned vector field is driven to average over all target velocities consistent with $(x_t, t, c)$, rather than selecting a single sub-mode-specific transport direction.

\paragraph{Dominant-mode bias.}
Suppose class $c$ contains $K$ sub-modes $\{S_k\}_{k=1}^K$. For a fixed $(x_t, t, c)$, let
\[
\pi_k(x_t, t, c) = p(x_1 \in S_k \mid x_t, t, c)
\]
denote the posterior weight of sub-mode $S_k$, where $\sum_{k=1}^K \pi_k(x_t, t, c)=1$. By the law of total expectation, Eq. \eqref{eq:optimal} can be written as
\begin{equation}
  v_\theta^*(x_t, t, c) = \sum_{k=1}^{K} \pi_k(x_t, t, c)\,
  \mathbb{E}[x_1 - x_0 \mid x_t, t, c,\, x_1 \in S_k].
  \label{eq:weighted}
\end{equation}
When multiple sub-modes remain possible for the same $(x_t, t, c)$, Eq. \eqref{eq:weighted} becomes a posterior-weighted aggregation of their mode-specific transport directions.
Consequently, dominant sub-modes contribute more strongly to the optimal vector field, whereas rare sub-modes contribute only weakly. This weighted combination across sub-modes is the essence of \emph{averaging distortion}. Because dominant sub-modes receive larger weights, the resulting vector field is biased toward them, which in turn induces a \emph{dominant-mode bias} in generation and reduces coverage of rare but valid variations.

\paragraph{Averaging distortion persists across step counts.}
Averaging distortion is not unique to one-step generation. Even when ODE integration is performed with multiple steps, the learned vector field is still obtained under the same class-conditional objective and may therefore remain biased toward dominant sub-modes.
In practice, samples are generated by numerically integrating the learned ODE (Eq.~\eqref{eq:ode}) with a fixed-step solver such as the Euler method. Using more integration steps (i.e., a smaller step size) reduces the discretization error of the numerical solver and produces a trajectory that more faithfully follows the learned vector field.
However, if the vector field itself is biased, such refinement only makes the solver trace the biased field more accurately, offering limited benefit for diversity recovery and potentially making the dominant-mode concentration more severe.
Our experiments (see \cref{sec:ablation}) are consistent with this interpretation: increasing the number of function evaluations (NFE) could improve sample quality (FID), while Recall remains flat or decreases slightly. These results suggest that the dominant-mode bias is rooted in the learned vector field itself, rather than in the discretization error of the ODE solver.

\subsection{Sub-mode Conditioned Flow Matching (SubFlow)}
\label{sec:subflow}

SubFlow mitigates averaging distortion by introducing a sub-mode index $k$ as an additional conditioning variable.

\paragraph{Sub-mode conditioned vector field.}
By conditioning on both the class label $c$ and a sub-mode index $k \in \{1, \ldots, K\}$, the conditional sub-distribution indexed by $(c,k)$ is expected to be substantially more concentrated and closer to unimodal.
The optimal predictor under SubFlow training is:
\begin{equation}
  v_\theta^*(x_t, t, c, k) = \mathbb{E}[x_1 - x_0 \mid x_t, t, c, k],
  \label{eq:subflow-optimal}
\end{equation}
which corresponds to the transport direction of the selected sub-mode.
Unlike Eq. \eqref{eq:weighted}, 
there is no averaging across different sub-modes under the same condition. \cref{fig:method} illustrates the complete SubFlow pipeline.

\paragraph{Sub-mode Discovery.}
We assign each training sample $x_1$ a sub-mode index $k$ via offline semantic clustering. Specifically, we first extract features $\phi(x_1)$ using a pre-trained DINOv3~\cite{simeoni2025dinov3} encoder, which provides semantically meaningful representations without requiring task-specific supervision. Then, for each class $c$, we partition the feature set $\{\phi(x_1^i)\}_{x_1^i \in c}$ into $K$ clusters and use the resulting cluster identity as the sub-mode label. In our implementation, we use $K$-Means and obtain centroids $\{\mu_k\}_{k=1}^K$, with assignments defined by
\begin{equation}
  k_i = \arg\min_{k \in \{1, \ldots, K\}} \|\phi(x_1^i) - \mu_k\|^2.
  \label{eq:assign}
\end{equation}
This procedure is performed once before training, and the resulting assignments $\{k_i\}$ augment the original class labels with no manual annotation cost. Importantly, SubFlow is not restricted to $K$-Means, and any clustering algorithm that produces a meaningful within-class partition can be used to define sub-mode labels, such as hierarchical clustering, Gaussian mixture models, or spectral clustering.

\paragraph{Training objective.}
The core idea of SubFlow is to extend the conditioning variables from $c$ to $(c,k)$, which can be applied to any flow-matching-based training objective. As a concrete example, integrating SubFlow into the standard conditional flow matching loss (Eq.~\eqref{eq:cfm}) yields:
\begin{equation}
  \mathcal{L}_{\mathrm{SubFlow}} = \mathbb{E}_{t,\, x_0,\, (x_1, c, k)} \bigl\| v_\theta(x_t, t, c, k) - (x_1 - x_0) \bigr\|^2,
  \label{eq:subflow-loss}
\end{equation}
where $k = k_i$ is the pre-assigned sub-mode label for sample $x_1^i$. The same principle applies when integrating SubFlow into other objectives such as MeanFlow or Shortcut Models by simply adding $k$ as an additional conditioning variable alongside $c$.

\begin{algorithm}[t]
\caption{SubFlow: Training and Inference}
\label{alg:subflow}
\begin{algorithmic}[1]
\Statex \textbf{Pre-processing (offline, once):}
\For{each class $c$}
  \State Extract semantic features $\phi(x_1^i)$ for samples in class $c$
  \State Partition them into $K$ sub-modes and store sub-mode indices $\{k_i\}$
  \State Estimate the empirical prior $p(k \mid c)$
\EndFor
\Statex
\Statex \textbf{Training:}
\For{each training iteration}
  \State Sample $(x_1, c, k)$ from the training set with sub-mode labels
  \State Sample $x_0 \sim p_0$, $t \sim \mathcal{U}[0,1]$; compute $x_t = (1{-}t)x_0 + tx_1$
  \State With probability $p_{\mathrm{drop}}$, set $c \leftarrow \varnothing$ \Comment{drop $c$ only; always retain $k$}
  \State Update $\theta$ by minimizing $\|v_\theta(x_t, t, c, k) - (x_1 - x_0)\|^2$
\EndFor
\Statex
\Statex \textbf{Inference (for a target class $c$):}
\State Sample $k \sim p(k \mid c)$, $\;x_0 \sim p_0$
\State Set $x_1 \leftarrow x_0 + v_{\mathrm{cfg}}(x_0, 0, c, k)$
\Comment{one-step Euler generation}
\State \Return $x_1$
\end{algorithmic}
\end{algorithm}

\begin{table*}[t]
\centering
\resizebox{0.78\textwidth}{!}{%
\begin{tabular}{l c c c c c c}
\toprule
Method & Param & Epoch & NFE & FID$\downarrow$ & Precision (\%) & Recall (\%) \\
\hline
\rowcolor{gray!20} \multicolumn{7}{l}{\textbf{GANs}} \\
BigGAN~\cite{brock2018large} {\scriptsize\color{gray}[ICLR'19]} & 112M & - & 1 & 7.70 & 82.17 & 27.67 \\
StyleGAN-XL~\cite{sauer2022stylegan} {\scriptsize\color{gray}[SIGGRAPH'22]} & 166M & 500 & 1 & 2.30 & 71.93 & 48.71 \\
GigaGAN~\cite{kang2023scaling} {\scriptsize\color{gray}[CVPR'23]} & 569M & 128 & 1 & 3.45 & 75.57 & 48.36 \\
\hline
\rowcolor{gray!20} \multicolumn{7}{l}{\textbf{Multi-step Diffusion/Flows}} \\
ADM~\cite{dhariwal2021diffusion} {\scriptsize\color{gray}[NeurIPS'21]} & 554M & 400 & 250 & 4.59 & 82 & 52 \\
DiT-XL/2~\cite{peebles2023scalable} {\scriptsize\color{gray}[ICCV'23]} & 676M & 1400 & 250 & 2.27 & 83 & 57 \\
SiT-XL/2~\cite{ma2024sit} {\scriptsize\color{gray}[ECCV'24]} & 675M & 1400 & 250 & 2.06 & 83 & 59 \\
SiT-XL/2+REPA~\cite{yu2025representation} {\scriptsize\color{gray}[ICLR'25]} & 675M & 800 & 250 & 1.42 & 80 & 65 \\
LightningDiT-XL/1~\cite{yao2025reconstruction} {\scriptsize\color{gray}[CVPR'25]} & 676M & 800 & 250 & 1.35 & 79 & 65 \\
DDT-XL/2~\cite{wang2025ddt} {\scriptsize\color{gray}[arXiv'25]} & 676M & 400 & 250 & 1.26 & 79 & 65 \\
RAE-XL/1~\cite{zheng2026diffusion} {\scriptsize\color{gray}[ICLR'26]} & 676M & 800 & 50 & 1.41 & 80 & 63 \\
\hline
\rowcolor{gray!20} \multicolumn{7}{l}{\textbf{Single-step Diffusion/Flows}} \\
AdversarialFlow-B/2~\cite{lin2025adversarial} {\scriptsize\color{gray}[arXiv'25]} & 131M & 200 & 1 & 3.05 & 73.37 & 50.35 \\
AlphaFlow-B/2~\cite{zhang2025alphaflow} {\scriptsize\color{gray}[arXiv'25]} & 131M & 240 & 1 & 5.40 & 69.94 & 43.52 \\
AlphaFlow-XL/2+~\cite{zhang2025alphaflow} {\scriptsize\color{gray}[arXiv'25]} & 676M & 240+60 & 1 & 2.58 & 63.97 & 58.85 \\
TiM-XL/2~\cite{wang2025transition} {\scriptsize\color{gray}[arXiv'25]} & 676M & 200 & 1 & 3.26 & 60.51 & 66.28 \\
pMF-B/16~\cite{lu2026one} {\scriptsize\color{gray}[arXiv'26]} & 131M & 320 & 1 & 3.12 & 72.30 & 50.53 \\
Shortcut-B/2~\cite{frans2025one} {\scriptsize\color{gray}[ICLR'25]} & 131M & 160 & 1 & 40.3 & 50.09 & 37.05 \\
\rowcolor{blue!8} Shortcut-B/2+SubFlow & 131M & 160 & 1 & \textbf{38.2} & 47.45 & \textbf{39.36} \\
MeanFlow-B/2~\cite{geng2025mean} {\scriptsize\color{gray}[NeurIPS'25]} & 131M & 240 & 1 & 6.17 & 70.35 & 43.45 \\
\rowcolor{blue!8} MeanFlow-B/2+SubFlow & 131M & 240 & 1 & \textbf{5.86} & 69.04 & \textbf{48.84} \\
SoFlow-B/2~\cite{luo2026soflow} {\scriptsize\color{gray}[ICLR'26]} & 131M & 240 & 1 & 4.85 & 70.99 & 47.78 \\
\rowcolor{blue!8} SoFlow-B/2+SubFlow & 131M & 240 & 1 & \textbf{4.26} & \textbf{71.45} & \textbf{49.43} \\
\bottomrule
\end{tabular}%
}
\vspace{8px}
\caption{Class-conditional ImageNet-256 generation results. NFE denotes the number of function evaluations (sampling steps). Results in the \textbf{Multi-step Diffusion/Flows} section are cited from \cite{zheng2026diffusion}, and all other results are evaluated using officially released checkpoints. SubFlow consistently improves Recall (diversity) when integrated into different one-step baselines, while maintaining competitive FID and Precision.}
\label{tab:main}
\end{table*}

\paragraph{Classifier-Free Guidance.}
For classifier-free guidance (CFG)~\cite{ho2022classifier}, we randomly replace the class label $c$ with a null token $\varnothing$ with probability $p_{\mathrm{drop}}$ during training, while always retaining the sub-mode index $k$. The reason is that, in SubFlow, $k$ is not used as an additional semantic condition to be guided; rather, it indexes which sub-mode-specific vector field is being learned within class $c$. Therefore, CFG should act only on the class condition, while keeping the selected sub-mode fixed.

Accordingly, the guided vector field is defined as
\begin{equation}
  v_{\mathrm{cfg}}(x_t, t, c, k)
  =
  v_\theta(x_t, t, \varnothing, k)
  + w \cdot \bigl(
  v_\theta(x_t, t, c, k)
  - v_\theta(x_t, t, \varnothing, k)
  \bigr),
  \label{eq:cfg}
\end{equation}
where $w \geq 1$ is the guidance scale. In practice, SubFlow is compatible with different inference-time guidance variants used by the underlying baseline, as long as the same sub-mode index $k$ is retained in both branches.

\paragraph{Inference.}
To preserve the empirical sub-mode frequency within each class at generation time, given a target class $c$, we first sample a sub-mode index $k$ from the empirical prior $p(k \mid c) = n_{c,k} / n_c$, where $n_{c,k}$ is the number of training samples in class $c$ assigned to sub-mode $k$ and $n_c$ is the total number of samples in class $c$. We then draw $x_0 \sim p_0$ and integrate the ODE using $v_{\mathrm{cfg}}(\cdot, \cdot, c, k)$. \Cref{alg:subflow} summarizes the complete procedure.

\subsection{Integration with Existing Methods}
\label{sec:integration}

SubFlow is \textit{plug-and-play}: integrating it into an existing one-step generation framework requires only three additions: (1)~offline semantic clustering to obtain sub-mode labels $\{k_i\}$ and per-class priors $p(k \mid c)$; (2)~a learned embedding for $k$ injected through the same conditioning pathway as the class label; and (3)~sampling $k \sim p(k \mid c)$ at inference. The backbone architecture, ODE solver, and optimization pipeline of the base method remain entirely unchanged. In particular, SubFlow does not modify the network architecture or introduce additional loss terms. Instead, it only augments the conditioning inputs with a finer-grained sub-mode signal.

\section{Experiments}
\label{sec:experiments}

We evaluate SubFlow on class-conditional ImageNet-256 generation. We first describe the experimental setup (\cref{sec:setup}), then present main quantitative comparisons against existing methods (\cref{sec:main-results}), followed by qualitative visualizations (\cref{sec:qualitative}) and ablation studies on key design choices (\cref{sec:ablation}).

\subsection{Experimental Setup}
\label{sec:setup}

\paragraph{Dataset and evaluation.}
We evaluate SubFlow on the ImageNet-256 benchmark~\cite{deng2009imagenet}, which contains 1,000 object classes with approximately 1.28 million training images at $256 \times 256$ resolution. We report Fr\'{e}chet Inception Distance (FID)~\cite{heusel2017gans} to measure overall image quality, and Precision/Recall~\cite{kynkaanniemi2019improved} to separately quantify fidelity and diversity. All metrics are computed against the full ImageNet training set using 50,000 generated samples.

\begin{figure*}[t]
  \centering
  \includegraphics[width=0.85\textwidth,trim=0 80 90 0,clip]{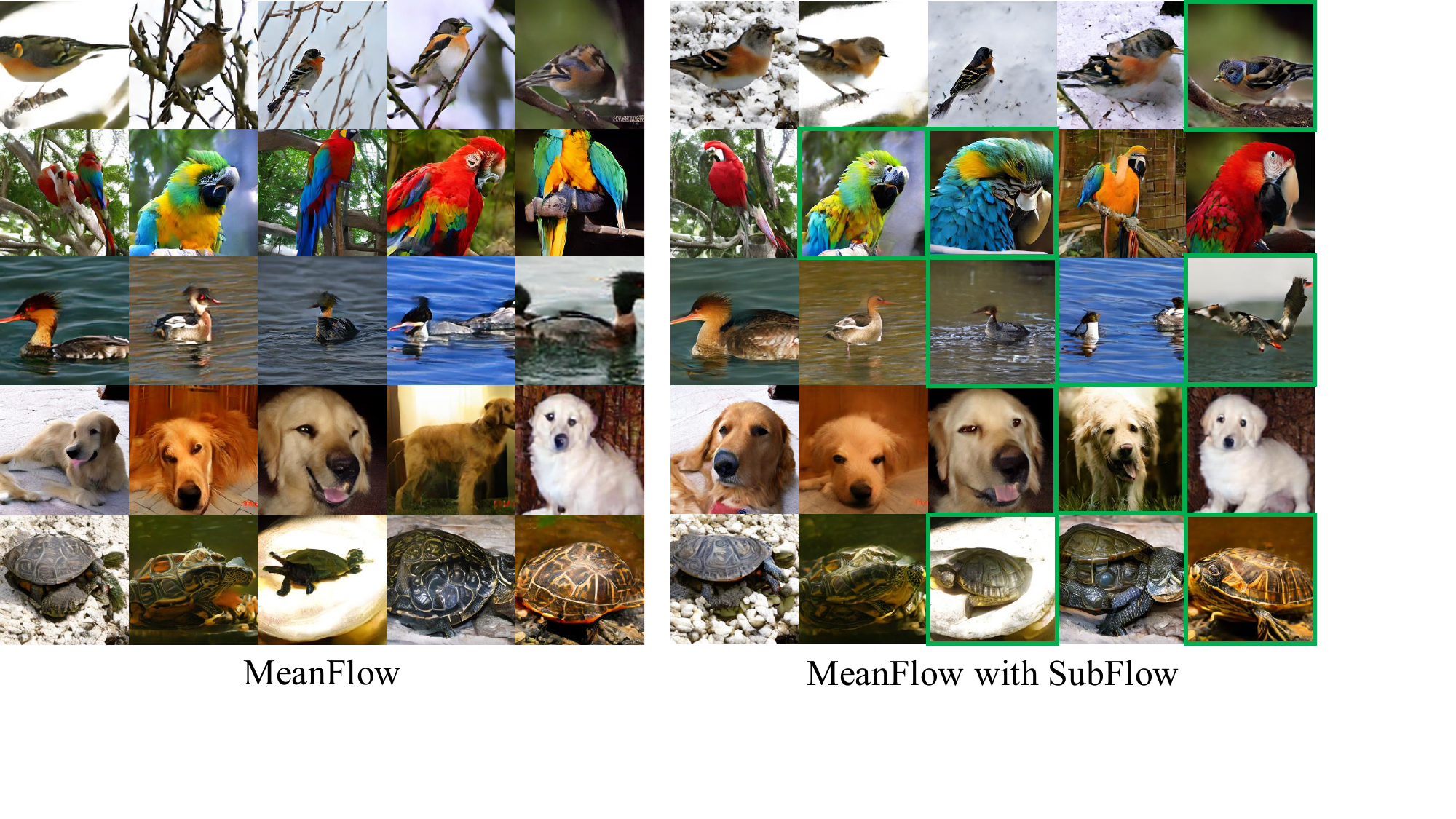}
  \vspace{-5px}
  \caption{Qualitative comparison between MeanFlow (left) and MeanFlow with SubFlow (right) on ImageNet-256. Each row shows five samples from the same class. Green boxes highlight samples where SubFlow produces visibly higher image quality with sharper details and fewer artifacts.}
  \label{fig:qualitative}
\end{figure*}

\paragraph{Integrated frameworks.}
To demonstrate the plug-and-play generality of SubFlow, we integrate it into three representative one-step generation frameworks:
\textbf{MeanFlow}~\cite{geng2025mean}, which learns a mean velocity field for direct one-step transport;
\textbf{Shortcut Models}~\cite{frans2025one}, which use self-consistency constraints to support variable-step generation;
and \textbf{SoFlow}~\cite{luo2026soflow}, which incorporates second-order supervision to improve sample fidelity.
For each framework, we compare the original model against the same model augmented with SubFlow conditioning.

\begin{table}[t]
\centering
\caption{Effect of the number of sampling steps (NFE) on MeanFlow-B/2 with and without SubFlow ($K{=}5$). Multi-step integration could improve FID but Recall slightly decreases, confirming that reduced discretization error amplifies rather than corrects the dominant-mode bias. SubFlow provides consistent Recall gains at all step counts.}
\vspace{-5px}
\label{tab:steps}
\resizebox{0.42\textwidth}{!}{
\begin{tabular}{l c c c c}
\toprule
Method & NFE & FID$\downarrow$ & Precision (\%) & Recall (\%) \\
\midrule
MeanFlow       & 1   & 6.17 & 70.35 & 43.45 \\
MeanFlow       & 2   & 5.46 & 80.82 & 38.91 \\
MeanFlow       & 4   & 5.54 & 82.86 & 37.59 \\
MeanFlow       & 8   & 5.58 & 83.35 & 37.62 \\
MeanFlow       & 16  & 5.60 & 83.27 & 37.83 \\
MeanFlow       & 32  & 5.64 & 83.12 & 38.16 \\
MeanFlow       & 64  & 5.66 & 83.00 & 37.78 \\
MeanFlow       & 128 & 5.69 & 82.81 & 38.02 \\
\midrule
+ SubFlow      & 1   & 5.86 & 69.04 & 48.84 \\
+ SubFlow      & 2   & 4.21 & 77.18 & 46.86 \\
+ SubFlow      & 4   & 3.92 & 79.17 & 45.88 \\
+ SubFlow      & 8   & 3.89 & 79.82 & 45.25 \\
+ SubFlow      & 16  & 3.90 & 79.71 & 45.43 \\
+ SubFlow      & 32  & 3.93 & 79.49 & 45.57 \\
+ SubFlow      & 64  & 3.95 & 79.48 & 45.54 \\
+ SubFlow      & 128 & 3.97 & 79.39 & 45.73 \\
\bottomrule
\end{tabular}
}
\vspace{-5px}
\end{table}

\paragraph{Implementation details.}

All experiments use a DiT-B/2 backbone \cite{peebles2023scalable} and follow the training configurations of their respective baseline methods to ensure a fair comparison. Specifically, MeanFlow is trained for 240 epochs ($\sim$1.2M steps) with Adam ($\beta_1{=}0.9$, $\beta_2{=}0.95$); Shortcut Models for 160 epochs ($\sim$800K steps) with AdamW ($\beta_1{=}0.9$, $\beta_2{=}0.999$, weight decay 0.1); and SoFlow for 240 epochs ($\sim$1.2M steps) with AdamW ($\beta_1{=}0.9$, $\beta_2{=}0.99$). All three use a learning rate of $1\times10^{-4}$, batch size 256, EMA, and a CFG drop rate of 0.1. SubFlow adds no extra hyperparameters beyond these baseline settings.
For sub-mode discovery, we extract semantic features using a pre-trained DINOv3~\cite{simeoni2025dinov3} encoder and apply $K$-Means clustering within each class. The number of sub-modes is set to $K = 5$ for all main experiments unless otherwise stated.

\subsection{Main Results on ImageNet-256}
\label{sec:main-results}

\Cref{tab:main} presents a comprehensive comparison on class-conditional ImageNet-256 generation. We organize the results into three categories: GANs, multi-step diffusion/flow models, and single-step diffusion/flow models.

Integrating SubFlow yields consistent improvements across all three one-step baselines. For MeanFlow-B/2, SubFlow improves Recall from 43.45\% to 48.84\% (+5.39\%) while also reducing FID (6.17$\to$5.86). For SoFlow-B/2, SubFlow improves FID (4.85$\to$4.26), Precision (70.99\%$\to$71.45\%) and Recall (47.78\%$\to$49.43\%).
For Shortcut-B/2, SubFlow can yield performance gains in Recall (37.05\%$\to$39.36\%) and FID (40.3$\to$38.2). Notably, the Recall of MeanFlow-B/2+SubFlow (48.84\%) surpasses that of several single-step methods with comparable or larger model sizes, such as AdversarialFlow-B/2 (50.35\% Recall but with adversarial training) and AlphaFlow-B/2 (43.52\%). Even compared with multi-step methods, MeanFlow-B/2+SubFlow achieves competitive diversity using only a single function evaluation: its Recall (48.84\%) is close to ADM (52\%) which requires 250 NFE with a 554M model. These results confirm that SubFlow provides a simple and effective way to improve diversity without sacrificing generation quality.

\paragraph{Multi-step generation does not recover diversity.}
As discussed in \cref{sec:avgdist}, averaging distortion is a structural property of the learned vector field, and it reduces discretization error through multi-step integration but does not correct this bias.
\cref{tab:steps} confirms this empirically using Euler integration: for MeanFlow, increasing NFE from 2 to 128 leads to a slight degradation in FID, and Recall stays around 38\%, indicating that reduced discretization error does not recover lost diversity. With SubFlow, diversity is substantially higher at all step counts (Recall $\approx$45\%--47\%), and quality also improves (FID 4.21$\to$3.97) when increasing NFE from 2 to 128. Notably, MeanFlow+SubFlow at NFE=2 already surpasses MeanFlow at NFE=128 in both FID and Recall, demonstrating that sub-mode conditioning is far more effective than additional integration steps for addressing dominant-mode bias.

\begin{figure}[t]
  \centering
  \includegraphics[width=0.48\textwidth,trim=0 250 150 0,clip]{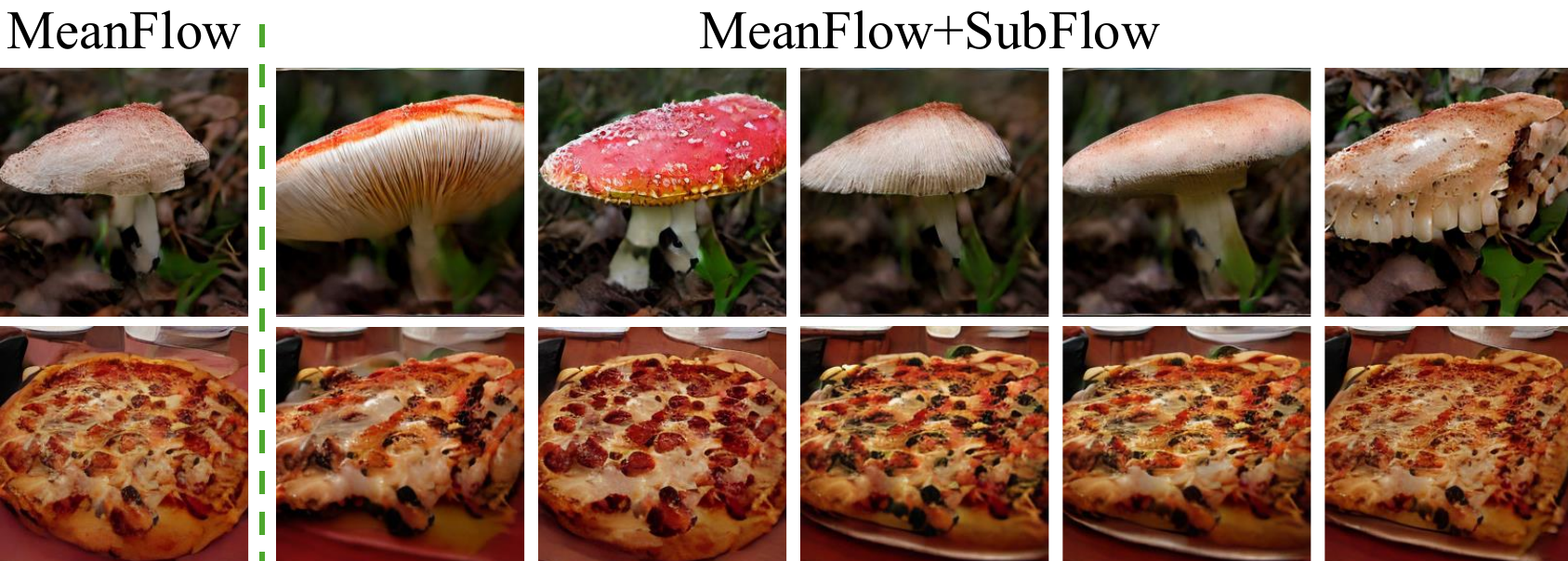}
  \caption{Diverse generation from the \emph{same noise} $x_0$. First column: MeanFlow output. Columns 2--6: MeanFlow+SubFlow with different sub-mode indices $k$.}
  \label{fig:diverse}
\end{figure}
\begin{figure}[t]
  \centering
  \begin{subfigure}[b]{0.49\linewidth}
    \includegraphics[width=\textwidth]{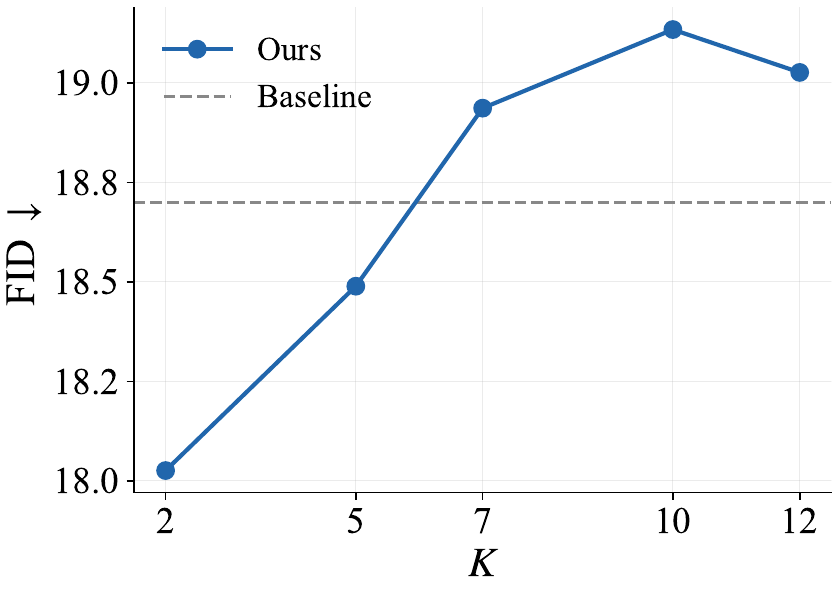}
    \vspace{-5mm}
    \caption{FID}
    \label{fig:sensitivity_K_fid}
  \end{subfigure}
  \begin{subfigure}[b]{0.49\linewidth}
    \includegraphics[width=\textwidth]{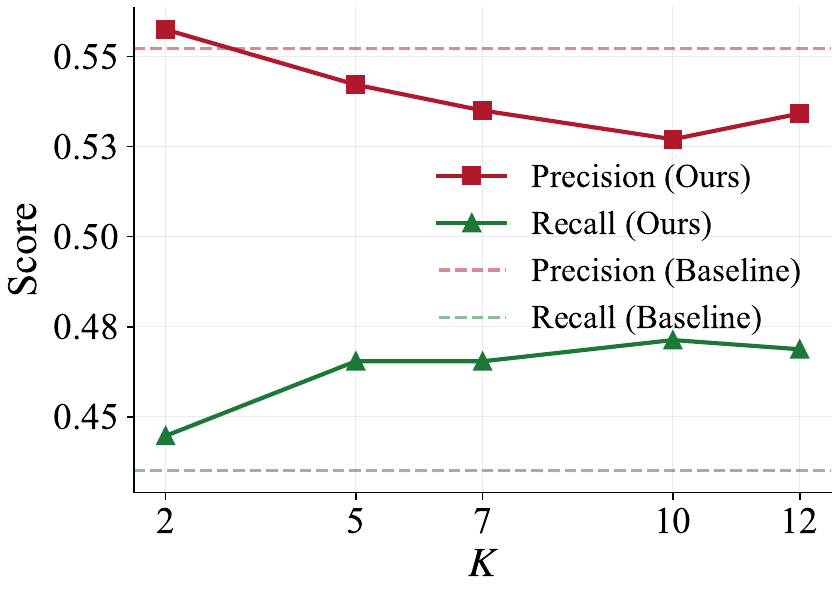}
    \vspace{-5mm}
    \caption{Precision / Recall}
    \label{fig:sensitivity_K_recall}
  \end{subfigure}
  \vspace{-5px}
  \caption{Sensitivity analysis for the number of sub-modes $K$. Dashed lines indicate the baseline without SubFlow. Recall improves steadily with moderate $K$, while Precision remains stable and FID stays competitive across a broad range.}
  \label{fig:sensitivity_K}
  \vspace{-10px}
\end{figure}

\subsection{Qualitative Results}
\label{sec:qualitative}

\cref{fig:qualitative} presents a side-by-side comparison of samples generated by MeanFlow and MeanFlow augmented with SubFlow for several ImageNet classes. For each class, five samples are shown per method. Samples highlighted with green boxes indicate cases where SubFlow produces visibly higher image quality, with sharper details, more coherent structures, and fewer artifacts compared to the baseline. This demonstrates that sub-mode conditioning not only improves diversity (as reflected by Recall in \cref{tab:main}) but also benefits per-sample visual fidelity, since the model no longer needs to compromise between conflicting transport directions from different sub-modes.

To further illustrate the diversity enabled by sub-mode conditioning, \cref{fig:diverse} shows samples generated from the \emph{same initial noise} $x_0$ but with different sub-mode indices $k$. The first column shows the single output produced by MeanFlow, which is deterministic given a fixed noise input. In contrast, columns 2--6 show the outputs of MeanFlow+SubFlow conditioned on five different sub-mode indices. Despite sharing the same noise, the generated images exhibit clearly distinct visual styles: for the mushroom class, different sub-mode indices produce variations in species, color, and shape; for the pizza class, the toppings, texture, and overall composition vary noticeably. This confirms that SubFlow successfully decomposes the intra-class distribution into semantically meaningful sub-modes, enabling diverse one-step generation from a single noise input.

\begin{table}[t]
\centering
\caption{Ablation on sub-mode clustering and sampling strategy (MeanFlow, DiT-B/4, 80 epochs, $K{=}5$). SubFlow ($K$-Means clustering and prior-based sampling $p(k \mid c)$) achieves a better quality--diversity trade-off than random sub-mode assignment and uniform sampling.}
\vspace{-5px}
\label{tab:sampling}
\begin{tabular}{l c c c}
\toprule
Strategy & FID$\downarrow$ & Prec.(\%) & Recall(\%) \\
\midrule
Baseline (no SubFlow) & 18.70 & 55.23 & 43.51 \\
Random sub-mode assignment & 18.80 & 55.45 & 43.92 \\
Uniform sampling & 20.00 & 52.86 & \textbf{48.22} \\
SubFlow & \textbf{18.49} & 54.21 & \textbf{46.54} \\
\bottomrule
\end{tabular}
\vspace{-10px}
\end{table}

\subsection{Ablation Studies}
\label{sec:ablation}

We conduct ablation experiments to analyze the key design choices of SubFlow, including the sensitivity to the number of sub-modes $K$ and the effectiveness of the clustering and sampling strategies. All ablations use MeanFlow with a DiT-B/4 backbone trained for 80 epochs on ImageNet-256.

\paragraph{Number of sub-modes $K$.}
The number of sub-modes $K$ is the key hyperparameter of SubFlow. \cref{fig:sensitivity_K} shows how FID, Precision, and Recall vary as $K$ ranges over $\{2, 5, 7, 10, 12\}$, with the corresponding baseline (without SubFlow) indicated by dashed lines. As $K$ increases from 2 to 5, Recall improves notably (44.5\%$\to$46.5\%) while FID remains close to the baseline. However, beyond $K=5$, further increasing $K$ yields diminishing returns in Recall but continues to degrade both FID and Precision. This is expected, as overly fine partitioning fragments semantically relevant modes into noisy sub-clusters, hurting generation quality without meaningful diversity gains. Based on this trade-off, we select $K=5$ for all main experiments, as it achieves the best balance between diversity improvement and quality preservation.

\paragraph{Sub-mode clustering and sampling strategy.}
\cref{tab:sampling} reports ablations on the sub-mode clustering and sampling strategy using the same DiT-B/4 setup at $K{=}5$. First, to verify that the semantically clustered labels are informative, we replace them with random sub-mode assignments, where each training sample receives a uniformly random $k \in \{1,\ldots,K\}$ independent of its semantics. Random assignment performs on par with the baseline (FID 18.80, Recall 43.92\%), confirming that the diversity gains of SubFlow stem from semantically meaningful partitions rather than merely increasing the number of conditioning indices.
Second, we study how the choice of inference-time sampling strategies for $k$ affects performance, comparing uniform sampling against the empirical prior $p(k \mid c)$.
Uniform sampling yields the highest Recall (48.22\%) but substantially degrades FID (20.00 vs.\ 18.49), likely because it over-samples rare sub-modes. In contrast, sampling from the empirical prior $p(k \mid c)$ achieves a more favorable quality--diversity trade-off (FID 18.49, Recall 46.54\%), and is therefore used in all experiments.

\section{Conclusion}

We identified \emph{averaging distortion}, which refers to the tendency of class-conditional flow models to learn a frequency-weighted mean over intra-class sub-modes. This phenomenon causes diversity degradation in few-step generation, and we show that increasing integration steps can not eliminate this bias.
To address this, we proposed SubFlow, which first partitions each class into semantically meaningful sub-modes via offline clustering on pre-trained features, and then conditions the vector field on the sub-mode index, so that each conditional sub-distribution is approximately unimodal and the learned flow accurately targets individual modes.
SubFlow can be easily integrated into existing one-step frameworks such as MeanFlow, Shortcut Models, and SoFlow. Experiments on ImageNet-256 demonstrate consistent improvements in Recall across all three baselines while maintaining competitive FID and Precision, validating SubFlow as a simple yet effective solution for restoring diversity in few-step flow matching.


%
%
\bibliographystyle{ACM-Reference-Format}
\bibliography{main}

@String(CVPR  = {IEEE Conf. Comput. Vis. Pattern Recog.})

@String(CVPR  = {CVPR})

@article{ho2020denoising,
  title={Denoising diffusion probabilistic models},
  author={Ho, Jonathan and Jain, Ajay and Abbeel, Pieter},
  journal={Advances in neural information processing systems},
  volume={33},
  pages={6840--6851},
  year={2020}
}

@article{song2020score,
  title={Score-based generative modeling through stochastic differential equations},
  author={Song, Yang and Sohl-Dickstein, Jascha and Kingma, Diederik P and Kumar, Abhishek and Ermon, Stefano and Poole, Ben},
  journal={arXiv preprint arXiv:2011.13456},
  year={2020}
}

@article{dhariwal2021diffusion,
  title={Diffusion models beat gans on image synthesis},
  author={Dhariwal, Prafulla and Nichol, Alexander},
  journal={Advances in neural information processing systems},
  volume={34},
  pages={8780--8794},
  year={2021}
}

@article{lipman2022flow,
  title={Flow matching for generative modeling},
  author={Lipman, Yaron and Chen, Ricky TQ and Ben-Hamu, Heli and Nickel, Maximilian and Le, Matt},
  journal={arXiv preprint arXiv:2210.02747},
  year={2022}
}

@article{tong2023improving,
  title={Improving and generalizing flow-based generative models with minibatch optimal transport},
  author={Tong, Alexander and Fatras, Kilian and Malkin, Nikolay and Huguet, Guillaume and Zhang, Yanlei and Rector-Brooks, Jarrid and Wolf, Guy and Bengio, Yoshua},
  journal={arXiv preprint arXiv:2302.00482},
  year={2023}
}

@article{chen2018neural,
  title={Neural ordinary differential equations},
  author={Chen, Ricky TQ and Rubanova, Yulia and Bettencourt, Jesse and Duvenaud, David K},
  journal={Advances in neural information processing systems},
  volume={31},
  year={2018}
}

@article{albergo2022building,
  title={Building normalizing flows with stochastic interpolants},
  author={Albergo, Michael S and Vanden-Eijnden, Eric},
  journal={arXiv preprint arXiv:2209.15571},
  year={2022}
}

@inproceedings{esser2024scaling,
  title={Scaling rectified flow transformers for high-resolution image synthesis},
  author={Esser, Patrick and Kulal, Sumith and Blattmann, Andreas and Entezari, Rahim and M{\"u}ller, Jonas and Saini, Harry and Levi, Yam and Lorenz, Dominik and Sauer, Axel and Boesel, Frederic and others},
  booktitle={Forty-first international conference on machine learning},
  year={2024}
}

@article{salimans2022progressive,
  title={Progressive distillation for fast sampling of diffusion models},
  author={Salimans, Tim and Ho, Jonathan},
  journal={arXiv preprint arXiv:2202.00512},
  year={2022}
}

@article{song2023consistency,
  title={Consistency models},
  author={Song, Yang and Dhariwal, Prafulla and Chen, Mark and Sutskever, Ilya},
  year={2023}
}

@inproceedings{meng2023distillation,
  title={On distillation of guided diffusion models},
  author={Meng, Chenlin and Rombach, Robin and Gao, Ruiqi and Kingma, Diederik and Ermon, Stefano and Ho, Jonathan and Salimans, Tim},
  booktitle={Proceedings of the IEEE/CVF conference on computer vision and pattern recognition},
  pages={14297--14306},
  year={2023}
}

@inproceedings{
liu2024instaflow,
title={InstaFlow: One Step is Enough for High-Quality Diffusion-Based Text-to-Image Generation},
author={Xingchao Liu and Xiwen Zhang and Jianzhu Ma and Jian Peng and qiang liu},
booktitle={The Twelfth International Conference on Learning Representations},
year={2024},
url={https://openreview.net/forum?id=1k4yZbbDqX}
}

@article{ho2022classifier,
  title={Classifier-free diffusion guidance},
  author={Ho, Jonathan and Salimans, Tim},
  journal={arXiv preprint arXiv:2207.12598},
  year={2022}
}

@article{kynkaanniemi2019improved,
  title={Improved precision and recall metric for assessing generative models},
  author={Kynk{\"a}{\"a}nniemi, Tuomas and Karras, Tero and Laine, Samuli and Lehtinen, Jaakko and Aila, Timo},
  journal={Advances in neural information processing systems},
  volume={32},
  year={2019}
}

@inproceedings{
brock2018large,
title={Large Scale {GAN} Training for High Fidelity Natural Image Synthesis},
author={Andrew Brock and Jeff Donahue and Karen Simonyan},
booktitle={International Conference on Learning Representations},
year={2019},
url={https://openreview.net/forum?id=B1xsqj09Fm},
}

@article{metz2016unrolled,
  title={Unrolled generative adversarial networks},
  author={Metz, Luke and Poole, Ben and Pfau, David and Sohl-Dickstein, Jascha},
  journal={arXiv preprint arXiv:1611.02163},
  year={2016}
}

@article{srivastava2017veegan,
  title={Veegan: Reducing mode collapse in gans using implicit variational learning},
  author={Srivastava, Akash and Valkov, Lazar and Russell, Chris and Gutmann, Michael U and Sutton, Charles},
  journal={Advances in neural information processing systems},
  volume={30},
  year={2017}
}

@article{shazeer2017outrageously,
  title={Outrageously large neural networks: The sparsely-gated mixture-of-experts layer},
  author={Shazeer, Noam and Mirhoseini, Azalia and Maziarz, Krzysztof and Davis, Andy and Le, Quoc and Hinton, Geoffrey and Dean, Jeff},
  journal={arXiv preprint arXiv:1701.06538},
  year={2017}
}

@article{van2017neural,
  title={Neural discrete representation learning},
  author={Van Den Oord, Aaron and Vinyals, Oriol and others},
  journal={Advances in neural information processing systems},
  volume={30},
  year={2017}
}

@inproceedings{
geng2025mean,
title={Mean Flows for One-step Generative Modeling},
author={Zhengyang Geng and Mingyang Deng and Xingjian Bai and J Zico Kolter and Kaiming He},
booktitle={The Thirty-ninth Annual Conference on Neural Information Processing Systems},
year={2025},
url={https://openreview.net/forum?id=uWj4s7rMnR}
}

@article{liu2022flow,
  title={Flow straight and fast: Learning to generate and transfer data with rectified flow},
  author={Liu, Xingchao and Gong, Chengyue and Liu, Qiang},
  journal={arXiv preprint arXiv:2209.03003},
  year={2022}
}

@inproceedings{
frans2025one,
title={One Step Diffusion via Shortcut Models},
author={Kevin Frans and Danijar Hafner and Sergey Levine and Pieter Abbeel},
booktitle={The Thirteenth International Conference on Learning Representations},
year={2025},
url={https://openreview.net/forum?id=OlzB6LnXcS}
}

@article{gandikota2025distilling,
  title={Distilling Diversity and Control in Diffusion Models},
  author={Gandikota, Rohit and Bau, David},
  journal={arXiv preprint arXiv:2503.10637},
  year={2025}
}

@article{simeoni2025dinov3,
  title={Dinov3},
  author={Sim{\'e}oni, Oriane and Vo, Huy V and Seitzer, Maximilian and Baldassarre, Federico and Oquab, Maxime and Jose, Cijo and Khalidov, Vasil and Szafraniec, Marc and Yi, Seungeun and Ramamonjisoa, Micha{\"e}l and others},
  journal={arXiv preprint arXiv:2508.10104},
  year={2025}
}

@article{lin2025adversarial,
  title={Adversarial Flow Models},
  author={Lin, Shanchuan and Yang, Ceyuan and Lin, Zhijie and Chen, Hao and Fan, Haoqi},
  journal={arXiv preprint arXiv:2511.22475},
  year={2025}
}

@inproceedings{deng2009imagenet,
  title={Imagenet: A large-scale hierarchical image database},
  author={Deng, Jia and Dong, Wei and Socher, Richard and Li, Li-Jia and Li, Kai and Fei-Fei, Li},
  booktitle={2009 IEEE conference on computer vision and pattern recognition},
  pages={248--255},
  year={2009},
  organization={Ieee}
}

@article{heusel2017gans,
  title={Gans trained by a two time-scale update rule converge to a local nash equilibrium},
  author={Heusel, Martin and Ramsauer, Hubert and Unterthiner, Thomas and Nessler, Bernhard and Hochreiter, Sepp},
  journal={Advances in neural information processing systems},
  volume={30},
  year={2017}
}

@article{zhang2025alphaflow,
  title={Alphaflow: Understanding and improving meanflow models},
  author={Zhang, Huijie and Siarohin, Aliaksandr and Menapace, Willi and Vasilkovsky, Michael and Tulyakov, Sergey and Qu, Qing and Skorokhodov, Ivan},
  journal={arXiv preprint arXiv:2510.20771},
  year={2025}
}

@inproceedings{sauer2022stylegan,
  title={Stylegan-xl: Scaling stylegan to large diverse datasets},
  author={Sauer, Axel and Schwarz, Katja and Geiger, Andreas},
  booktitle={ACM SIGGRAPH 2022 conference proceedings},
  pages={1--10},
  year={2022}
}

@inproceedings{kang2023scaling,
  title={Scaling up gans for text-to-image synthesis},
  author={Kang, Minguk and Zhu, Jun-Yan and Zhang, Richard and Park, Jaesik and Shechtman, Eli and Paris, Sylvain and Park, Taesung},
  booktitle={Proceedings of the IEEE/CVF conference on computer vision and pattern recognition},
  pages={10124--10134},
  year={2023}
}

@inproceedings{
zheng2026diffusion,
title={Diffusion Transformers with Representation Autoencoders},
author={Boyang Zheng and Nanye Ma and Shengbang Tong and Saining Xie},
booktitle={The Fourteenth International Conference on Learning Representations},
year={2026},
url={https://openreview.net/forum?id=0u1LigJaab}
}

@article{wang2025ddt,
  title={Ddt: Decoupled diffusion transformer},
  author={Wang, Shuai and Tian, Zhi and Huang, Weilin and Wang, Limin},
  journal={arXiv preprint arXiv:2504.05741},
  year={2025}
}

@inproceedings{ma2024sit,
  title={Sit: Exploring flow and diffusion-based generative models with scalable interpolant transformers},
  author={Ma, Nanye and Goldstein, Mark and Albergo, Michael S and Boffi, Nicholas M and Vanden-Eijnden, Eric and Xie, Saining},
  booktitle={European Conference on Computer Vision},
  pages={23--40},
  year={2024},
  organization={Springer}
}

@inproceedings{
yu2025representation,
title={Representation Alignment for Generation: Training Diffusion Transformers Is Easier Than You Think},
author={Sihyun Yu and Sangkyung Kwak and Huiwon Jang and Jongheon Jeong and Jonathan Huang and Jinwoo Shin and Saining Xie},
booktitle={The Thirteenth International Conference on Learning Representations},
year={2025},
url={https://openreview.net/forum?id=DJSZGGZYVi}
}

@inproceedings{yao2025reconstruction,
  title={Reconstruction vs. generation: Taming optimization dilemma in latent diffusion models},
  author={Yao, Jingfeng and Yang, Bin and Wang, Xinggang},
  booktitle={Proceedings of the Computer Vision and Pattern Recognition Conference},
  pages={15703--15712},
  year={2025}
}

@inproceedings{peebles2023scalable,
  title={Scalable diffusion models with transformers},
  author={Peebles, William and Xie, Saining},
  booktitle={Proceedings of the IEEE/CVF international conference on computer vision},
  pages={4195--4205},
  year={2023}
}

@article{wang2025transition,
  title={Transition models: Rethinking the generative learning objective},
  author={Wang, Zidong and Zhang, Yiyuan and Yue, Xiaoyu and Yue, Xiangyu and Li, Yangguang and Ouyang, Wanli and Bai, Lei},
  journal={arXiv preprint arXiv:2509.04394},
  year={2025}
}

@article{lin2025beyond,
  title={Beyond optimal transport: Model-aligned coupling for flow matching},
  author={Lin, Yexiong and Yao, Yu and Liu, Tongliang},
  journal={arXiv preprint arXiv:2505.23346},
  year={2025}
}

@inproceedings{cao2025force,
  title={Force matching with relativistic constraints: A physics-inspired approach to stable and efficient generative modeling},
  author={Cao, Yang and Chen, Bo and Li, Xiaoyu and Liang, Yingyu and Sha, Zhizhou and Shi, Zhenmei and Song, Zhao and Wan, Mingda},
  booktitle={Proceedings of the 34th ACM International Conference on Information and Knowledge Management},
  pages={179--188},
  year={2025}
}

@inproceedings{
zhang2025towards,
title={Towards Hierarchical Rectified Flow},
author={Yichi Zhang and Yici Yan and Alex Schwing and Zhizhen Zhao},
booktitle={The Thirteenth International Conference on Learning Representations},
year={2025},
url={https://openreview.net/forum?id=6F6qwdycgJ}
}

@inproceedings{su2025high,
  title={High-order flow matching: Unified framework and sharp statistical rates},
  author={Su, Maojiang and Hu, Jerry Yao-Chieh and Lee, Yi-Chen and Zhu, Ning and Chung, Jui-Hui and Wu, Shang and Song, Zhao and Chen, Minshuo and Liu, Han},
  booktitle={The Thirty-ninth Annual Conference on Neural Information Processing Systems},
  year={2025}
}

@article{lu2026one,
  title={One-step Latent-free Image Generation with Pixel Mean Flows},
  author={Lu, Yiyang and Lu, Susie and Sun, Qiao and Zhao, Hanhong and Jiang, Zhicheng and Wang, Xianbang and Li, Tianhong and Geng, Zhengyang and He, Kaiming},
  journal={arXiv preprint arXiv:2601.22158},
  year={2026}
}

@inproceedings{
luo2026soflow,
title={SoFlow: Solution Flow Models for One-Step Generative Modeling},
author={Tianze Luo and Haotian Yuan and Zhuang Liu},
booktitle={The Fourteenth International Conference on Learning Representations},
year={2026},
url={https://openreview.net/forum?id=cjb03GNqYw}
}

@article{song2019generative,
  title={Generative modeling by estimating gradients of the data distribution},
  author={Song, Yang and Ermon, Stefano},
  journal={Advances in neural information processing systems},
  volume={32},
  year={2019}
}

@article{karras2022elucidating,
  title={Elucidating the design space of diffusion-based generative models},
  author={Karras, Tero and Aittala, Miika and Aila, Timo and Laine, Samuli},
  journal={Advances in neural information processing systems},
  volume={35},
  pages={26565--26577},
  year={2022}
}

@inproceedings{rombach2022high,
  title={High-resolution image synthesis with latent diffusion models},
  author={Rombach, Robin and Blattmann, Andreas and Lorenz, Dominik and Esser, Patrick and Ommer, Bj{\"o}rn},
  booktitle={Proceedings of the IEEE/CVF conference on computer vision and pattern recognition},
  pages={10684--10695},
  year={2022}
}

@article{goodfellow2014generative,
  title={Generative adversarial nets},
  author={Goodfellow, Ian J and Pouget-Abadie, Jean and Mirza, Mehdi and Xu, Bing and Warde-Farley, David and Ozair, Sherjil and Courville, Aaron and Bengio, Yoshua},
  journal={Advances in neural information processing systems},
  volume={27},
  year={2014}
}

@inproceedings{yin2024one,
  title={One-step diffusion with distribution matching distillation},
  author={Yin, Tianwei and Gharbi, Micha{\"e}l and Zhang, Richard and Shechtman, Eli and Durand, Fredo and Freeman, William T and Park, Taesung},
  booktitle={Proceedings of the IEEE/CVF conference on computer vision and pattern recognition},
  pages={6613--6623},
  year={2024}
}

@article{luo2024one,
  title={One-step diffusion distillation through score implicit matching},
  author={Luo, Weijian and Huang, Zemin and Geng, Zhengyang and Kolter, J Zico and Qi, Guo-jun},
  journal={Advances in Neural Information Processing Systems},
  volume={37},
  pages={115377--115408},
  year={2024}
}

@inproceedings{zhou2024score,
  title={Score identity distillation: Exponentially fast distillation of pretrained diffusion models for one-step generation},
  author={Zhou, Mingyuan and Zheng, Huangjie and Wang, Zhendong and Yin, Mingzhang and Huang, Hai},
  booktitle={Forty-first International Conference on Machine Learning},
  year={2024}
}

@article{luo2023diff,
  title={Diff-instruct: A universal approach for transferring knowledge from pre-trained diffusion models},
  author={Luo, Weijian and Hu, Tianyang and Zhang, Shifeng and Sun, Jiacheng and Li, Zhenguo and Zhang, Zhihua},
  journal={Advances in Neural Information Processing Systems},
  volume={36},
  pages={76525--76546},
  year={2023}
}

@article{song2023improved,
  title={Improved techniques for training consistency models},
  author={Song, Yang and Dhariwal, Prafulla},
  journal={arXiv preprint arXiv:2310.14189},
  year={2023}
}

@inproceedings{
lu2025simplifying,
title={Simplifying, Stabilizing and Scaling Continuous-time Consistency Models},
author={Cheng Lu and Yang Song},
booktitle={The Thirteenth International Conference on Learning Representations},
year={2025},
url={https://openreview.net/forum?id=LyJi5ugyJx}
}

@article{geng2024consistency,
  title={Consistency models made easy},
  author={Geng, Zhengyang and Pokle, Ashwini and Luo, William and Lin, Justin and Kolter, J Zico},
  journal={arXiv preprint arXiv:2406.14548},
  year={2024}
}

@inproceedings{
kim2024consistency,
title={Consistency Trajectory Models: Learning Probability Flow {ODE} Trajectory of Diffusion},
author={Dongjun Kim and Chieh-Hsin Lai and Wei-Hsiang Liao and Naoki Murata and Yuhta Takida and Toshimitsu Uesaka and Yutong He and Yuki Mitsufuji and Stefano Ermon},
booktitle={The Twelfth International Conference on Learning Representations},
year={2024},
url={https://openreview.net/forum?id=ymjI8feDTD}
}

@inproceedings{
zhou2025inductive,
title={Inductive Moment Matching},
author={Linqi Zhou and Stefano Ermon and Jiaming Song},
booktitle={Forty-second International Conference on Machine Learning},
year={2025},
url={https://openreview.net/forum?id=pwNSUo7yUb}
}

@article{chen2025pi,
  title={pi-flow: Policy-based few-step generation via imitation distillation},
  author={Chen, Hansheng and Zhang, Kai and Tan, Hao and Guibas, Leonidas and Wetzstein, Gordon and Bi, Sai},
  journal={arXiv preprint arXiv:2510.14974},
  year={2025}
}

@inproceedings{ester1996density,
  title={A density-based algorithm for discovering clusters in large spatial databases with noise},
  author={Ester, Martin and Kriegel, Hans-Peter and Sander, J{\"o}rg and Xu, Xiaowei and others},
  booktitle={kdd},
  volume={96},
  number={34},
  pages={226--231},
  year={1996}
}

@article{kynkaanniemi2024applying,
  title={Applying guidance in a limited interval improves sample and distribution quality in diffusion models},
  author={Kynk{\"a}{\"a}nniemi, Tuomas and Aittala, Miika and Karras, Tero and Laine, Samuli and Aila, Timo and Lehtinen, Jaakko},
  journal={Advances in Neural Information Processing Systems},
  volume={37},
  pages={122458--122483},
  year={2024}
}

@inproceedings{
sadat2024cads,
title={{CADS}: Unleashing the Diversity of Diffusion Models through Condition-Annealed Sampling},
author={Seyedmorteza Sadat and Jakob Buhmann and Derek Bradley and Otmar Hilliges and Romann M. Weber},
booktitle={The Twelfth International Conference on Learning Representations},
year={2024},
url={https://openreview.net/forum?id=zMoNrajk2X}
}

@inproceedings{campello2013density,
  title={Density-based clustering based on hierarchical density estimates},
  author={Campello, Ricardo JGB and Moulavi, Davoud and Sander, J{\"o}rg},
  booktitle={Pacific-Asia conference on knowledge discovery and data mining},
  pages={160--172},
  year={2013},
  organization={Springer}
}

@inproceedings{qu2025spatialvla,
  title={Spatialvla: Exploring spatial representations for visual-language-action model},
  author={Qu, Delin and Song, Haoming and Chen, Qizhi and Yao, Yuanqi and Ye, Xinyi and Ding, Yan and Wang, Zhigang and Gu, JiaYuan and Zhao, Bin and Wang, Dong and others},
  journal={Robotics: Science and Systems},
  year={2025}
}

@article{qu2025eo,
  title={EO-1: Interleaved Vision-Text-Action Pretraining for General Robot Control},
  author={Qu, Delin and Song, Haoming and Chen, Qizhi and Chen, Zhaoqing and Gao, Xianqiang and Ye, Xinyi and Lv, Qi and Shi, Modi and Ren, Guanghui and Ruan, Cheng and others},
  journal={arXiv preprint arXiv:2508.21112},
  year={2025}
}

@article{xiang2026safety,
  title={When Safety Collides: Resolving Multi-Category Harmful Conflicts in Text-to-Image Diffusion via Adaptive Safety Guidance},
  author={Xiang, Yongli and Hong, Ziming and Wang, Zhaoqing and Zhao, Xiangyu and Han, Bo and Liu, Tongliang},
  journal={arXiv preprint arXiv:2602.20880},
  year={2026}
}

@article{zheng2026vii,
  title={VII: Visual Instruction Injection for Jailbreaking Image-to-Video Generation Models},
  author={Zheng, Bowen and Xiang, Yongli and Hong, Ziming and Lin, Zerong and Yu, Chaojian and Liu, Tongliang and You, Xinge},
  journal={arXiv preprint arXiv:2602.20999},
  year={2026}
}

@article{hong2025adlift,
  title={AdLift: Lifting Adversarial Perturbations to Safeguard 3D Gaussian Splatting Assets Against Instruction-Driven Editing},
  author={Hong, Ziming and Huang, Tianyu and Chen, Runnan and Ye, Shanshan and Gong, Mingming and Han, Bo and Liu, Tongliang},
  journal={arXiv preprint arXiv:2512.07247},
  year={2025}
}

@inproceedings{
zheng2025aligning,
title={Aligning What Matters: Masked Latent Adaptation for Text-to-Audio-Video Generation},
author={Jiyang Zheng and Siqi Pan and Yu Yao and Zhaoqing Wang and Dadong Wang and Tongliang Liu},
booktitle={The Thirty-ninth Annual Conference on Neural Information Processing Systems},
year={2025},
url={https://openreview.net/forum?id=I0hRN2HMeH}
}

@InProceedings{Zhou_2024_CVPR,
    author    = {Zhou, Yang and Shao, Hao and Wang, Letian and Waslander, Steven L. and Li, Hongsheng and Liu, Yu},
    title     = {SmartRefine: A Scenario-Adaptive Refinement Framework for Efficient Motion Prediction},
    booktitle = {Proceedings of the IEEE/CVF Conference on Computer Vision and Pattern Recognition (CVPR)},
    month     = {June},
    year      = {2024},
    pages     = {15281-15290}
}

@inproceedings{
zhou2026drivinggen,
title={DrivingGen: A Comprehensive Benchmark for Generative Video World Models in Autonomous Driving},
author={Yang Zhou and Hao Shao and Letian Wang and Zhuofan Zong and Hongsheng Li and Steven L. Waslander},
booktitle={The Fourteenth International Conference on Learning Representations},
year={2026},
url={https://openreview.net/forum?id=OrgL5DsU0f}
}

@misc{zhou2026drivedreamerpolicy,
      title={DriveDreamer-Policy: A Geometry-Grounded World-Action Model for Unified Generation and Planning}, 
      author={Yang Zhou and Xiaofeng Wang and Hao Shao and Letian Wang and Guosheng Zhao and Jiangnan Shao and Jiagang Zhu and Tingdong Yu and Zheng Zhu and Guan Huang and Steven L. Waslander},
      year={2026},
      eprint={2604.01765},
      archivePrefix={arXiv},
      primaryClass={cs.CV},
      url={https://arxiv.org/abs/2604.01765}, 
}

\newpage

\appendix

\section{Additional Ablation Studies}

All ablation experiments in this supplementary material use MeanFlow with a DiT-B/4 backbone trained for 80 epochs on ImageNet-256, with $K{=}5$ sub-modes, consistent with the ablation setup in the main paper.

\subsection{Alternative Clustering Algorithm}
\label{sec:suppl-dbscan}

In the main paper, we use $K$-Means clustering on DINOv3 features to discover sub-modes within each class. Here we investigate whether the choice of clustering algorithm significantly affects SubFlow's performance by replacing $K$-Means with DBSCAN~\cite{ester1996density}, a density-based clustering method that does not require a predefined number of clusters.

Unlike $K$-Means, which partitions the feature space into exactly $K$ Voronoi cells, DBSCAN groups points based on local density and can discover clusters of arbitrary shape while marking low-density points as noise. This makes it a natural alternative for sub-mode discovery, since intra-class structure may not always conform to spherical clusters.

\begin{table}[h]
\centering
\caption{Comparison of clustering algorithms for sub-mode discovery (MeanFlow, DiT-B/4, 80 epochs). $K$-Means yields better FID and comparable Recall, suggesting that evenly partitioned sub-modes are more effective for flow matching conditioning.}
\vspace{-5px}
\label{tab:clustering-algo}
\begin{tabular}{l c c c}
\toprule
Clustering Algorithm & FID$\downarrow$ & Prec.(\%) & Recall(\%) \\
\midrule
Baseline (no SubFlow) & 18.70 & 55.23 & 43.51 \\
DBSCAN & 19.49 & 54.65 & 44.67 \\
$K$-Means (SubFlow) & \textbf{18.49} & 54.21 & \textbf{46.54} \\
\bottomrule
\end{tabular}
\vspace{-5px}
\end{table}

\cref{tab:clustering-algo} compares the two clustering algorithms. DBSCAN improves Recall over the baseline (44.67\% vs.\ 43.51\%), confirming that density-based sub-mode discovery can also provide diversity gains. However, it underperforms $K$-Means in both FID (19.49 vs.\ 18.49) and Recall (44.67\% vs.\ 46.54\%). We attribute this gap to two factors. First, DBSCAN produces clusters of varying sizes and may leave some samples unassigned as noise, leading to incomplete sub-mode partitions. Second, the number of effective clusters discovered by DBSCAN is data-dependent and may not match the optimal granularity for each class.
In contrast, $K$-Means provides a consistent partition into exactly $K$ sub-modes across all classes, which aligns well with the fixed-dimensional sub-mode embedding used in SubFlow. These results suggest that while SubFlow is compatible with different clustering algorithms, $K$-Means offers a favorable combination of simplicity and effectiveness.

\subsection{Effect of Dropping Sub-mode Index During Training}
\label{sec:suppl-drop-k}

A key design choice in SubFlow is that during classifier-free guidance (CFG) training, only the class label $c$ is randomly dropped while the sub-mode index $k$ is always retained (see Section 3.3 of the main paper). Here we empirically validate this design by comparing it against the alternative of also randomly dropping $k$ during training.

\begin{table}[h]
\centering
\caption{Effect of dropping the sub-mode index $k$ during CFG training (MeanFlow, DiT-B/4, 80 epochs, $K{=}5$). Both $c$ and $k$ are dropped independently with the same probability. Dropping $c$ only (our default) achieves the best overall trade-off.}
\vspace{-5px}
\label{tab:drop-k}
\begin{tabular}{l c c c}
\toprule
Strategy & FID$\downarrow$ & Prec.(\%) & Recall(\%) \\
\midrule
Baseline (no SubFlow) & 18.70 & 55.23 & 43.51 \\
SubFlow (drop both $c$ and $k$) & 18.59 & 54.79 & 46.25 \\
SubFlow (drop $c$ only) & \textbf{18.49} & 54.21 & \textbf{46.54} \\
\bottomrule
\end{tabular}
\vspace{-5px}
\end{table}

\cref{tab:drop-k} shows the results. When both $c$ and $k$ are independently dropped during training, the performance is close to but slightly worse than dropping $c$ only (FID 18.59 vs.\ 18.49, Recall 46.25\% vs.\ 46.54\%). Although the gap is modest, the trend is consistent across all three metrics and supports our design rationale: $k$ serves as a structural index that selects which sub-mode-specific vector field is active, rather than a semantic condition to be guided. Dropping $k$ during training introduces an unconditional branch that averages over all sub-modes, partially reintroducing the averaging distortion that SubFlow aims to eliminate. Retaining $k$ in both branches ensures that CFG operates purely along the class-conditioning axis, preserving the sub-mode-specific structure of the learned vector field.

\section{Visualization of Intra-class Structure}
\label{sec:suppl-tsne}

\begin{figure*}[t]
  \centering
  \begin{subfigure}[t]{0.45\linewidth}
    \centering
    \includegraphics[width=\linewidth]{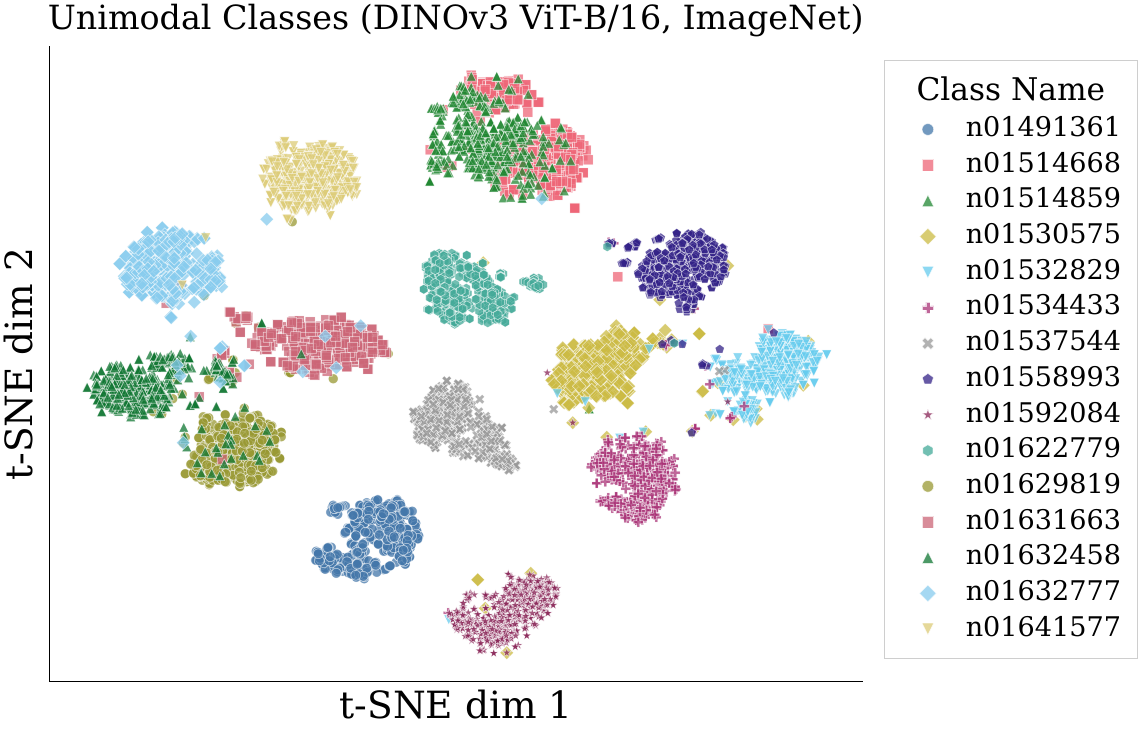}
    \caption{Unimodal classes (15 lowest clustering scores).
      Each class forms a single compact cluster.}
    \label{fig:tsne-uni}
  \end{subfigure}
  \begin{subfigure}[t]{0.45\linewidth}
    \centering
    \includegraphics[width=\linewidth]{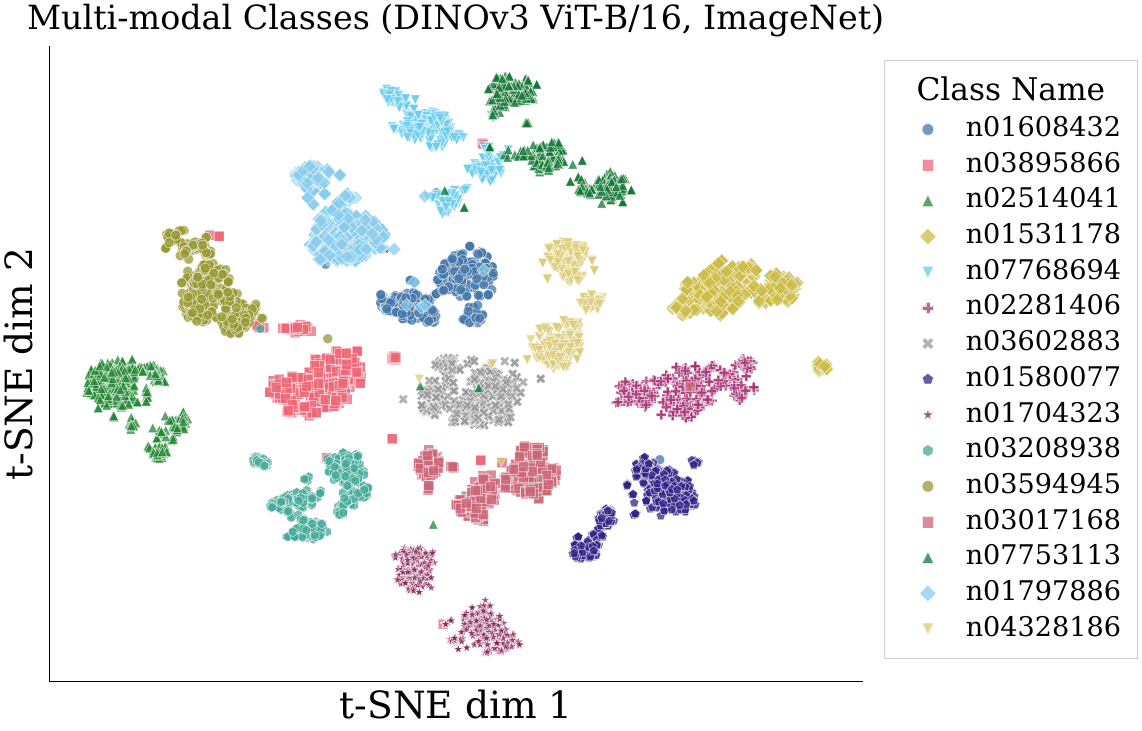}
    \caption{Multi-modal classes (15 highest clustering scores).
      Each class splits into 2--4 distinct sub-clusters.}
    \label{fig:tsne-multi}
  \end{subfigure}
  \caption{t-SNE visualization of DINOv3 features for ImageNet
    classes at opposite ends of the HDBSCAN clustering spectrum.
    Classes in~(a) are well served by standard class conditioning,
    while classes in~(b) suffer from averaging distortion that
    SubFlow's sub-mode conditioning resolves.}
  \label{fig:tsne}
\end{figure*}

To provide intuition for why sub-mode conditioning is beneficial,
we visualize the intra-class feature structure of ImageNet using
t-SNE applied to DINOv3 ViT-B/16 features.

We run HDBSCAN~\cite{campello2013density} independently on the features of each ImageNet class
and rank all 1{,}000 classes by a composite clustering score that combines
the number of discovered sub-clusters, silhouette coefficient,
and HDBSCAN persistence.
Out of the 1{,}000 classes, a substantial fraction exhibits
multi-modal structure with two or more stable sub-clusters,
while many others form a single compact cluster in feature space.

\Cref{fig:tsne} contrasts the two extremes.
The 15 lowest-scoring classes (\cref{fig:tsne-uni}) each collapse
into a single tight cluster, indicating that samples within the
class are semantically homogeneous.
For these classes, class-conditional flow matching already provides
an adequate conditioning signal.
In contrast, the 15 highest-scoring classes (\cref{fig:tsne-multi})
display two to four well-separated sub-clusters, corresponding to
distinct visual variations such as different subspecies, poses,
or backgrounds.
For these multi-modal classes, a single class-conditional vector
field must average over the distinct sub-modes, leading to the
averaging distortion analyzed in Section~3.2 of the main paper.

SubFlow addresses this by assigning each sub-cluster its own
conditioning index~$(c, k)$, so that the vector field within each
partition targets a locally unimodal distribution.
This explains why SubFlow's improvement concentrates on
Recall, recovering diverse modes that class-conditional generation
tends to collapse, rather than Precision, and why the overall FID
gain is driven primarily by the multi-modal subset of classes.

\end{document}